\documentclass[conference]{IEEEtran}
\IEEEoverridecommandlockouts
\usepackage{cite}
\usepackage{amsmath,amssymb,amsfonts}
\usepackage{algorithmic}
\usepackage{graphicx}
\usepackage{textcomp}
\usepackage{xcolor}
\usepackage{hyperref}
\hypersetup{
    colorlinks=true,
    linkcolor=blue,
    filecolor=magenta,      
    urlcolor=blue,
}
\usepackage[numbers,sort&compress]{natbib}
\def\BibTeX{{\rm B\kern-.05em{\sc i\kern-.025em b}\kern-.08em
    T\kern-.1667em\lower.7ex\hbox{E}\kern-.125emX}}

\usepackage{caption}
\usepackage{subfigure}
\usepackage{float}

\usepackage[T1]{fontenc}
\usepackage{fancyhdr}
\title{Predicting Blossom Date of Cherry Tree With Support Vector Machine and Recurrent Neural Network}
\author{\href{mailto:hz2212@nyu.edu}{Hongyi Zheng}, \href{mailto:yc3823@nyu.edu}{Yanyu Chen}, \href{mailto:zz2589@nyu.edu}{Zihan Zhang}}

\date{May 2021}

\begin{document}

\maketitle

\thispagestyle{plain}
\setcounter{page}{1}
\pagenumbering{arabic}

\begin{abstract}
    \noindent\textit{Our project probes the relationship between temperatures and the blossom date of cherry trees. Through modeling, future flowering will become predictive, helping the public plan travels and avoid pollen season. To predict the date when the cherry trees will blossom exactly could be viewed as a multiclass classification problem, so we applied the multi-class Support Vector Classifier (SVC) and Recurrent Neural Network (RNN), particularly Long Short-term Memory (LSTM), to formulate the problem. In the end, we evaluate and compare the performance of these approaches to find out which one might be more applicable in reality. }
\end{abstract}

\section{Introduction}
Many plants have high ornamental value during specific phenophases, and plant phenology correlates highly with seasonal vegetation landscape. Determination of the span and spatiotemporal patterns of the tourism season for ornamental plants could provide tourism administrators and the tourists themselves with a theoretical basis for making travel arrangements. Cherry Blossom, as our investigation focus, is widely distributed in the northern hemisphere, including Japan, China, and United States, and is tremendously important both culturally and economically. According to The National News review, during the 2018 hanami season, an estimated 63 million people travel to and within Japan (more than 40\% of foreign visitors) with total spending of around \$2.7 billion. 

Our primary objective is to implement ML techniques to predict the future exact peak blossom date of Cherry trees given past sequential daily temperature records (average, max, min, etc.). Such causality and correlation are inspired by Zhang's observation: among all meteorological features, daily average temperature correlates to the first flowering date and full flowering date of ornamental plants (Magnolia, Subhirtella) in the Beijing area most strongly \cite{jiwen_model}. The accumulated temperature of a consecutive time span describes the growing process of plants and if the value exceeds a certain threshold, the tree will blossom. Also, the paper suggests adding additional factors and features, like relative humidity, solar radiation, and wind speed measurement might improve the prediction accuracy. We identify the problem as a multi-class classification problem. We implement SVM to evaluate the non-sequential time interval and LSTM RNN to describe the interaction between different specific timestamps in sequential series. The effectiveness and universality of these two approaches in the temperature forecasting field are respectively examined and presented \cite{accumulated_temp, en13164215}. SVM is preferred based on its good compromise between simplicity and accuracy; An artificial neural network is more applicable than a regression model when predicting accumulated temperature. 

Our research field is novel and uncultivated after taking an online literature review. Present studies mostly apply thermal-time-based or process-based phenology models and statistical parameterizations, but the ML application is scarce \cite{cherry_tree, cherry_2, cherry_3}. For its sensitivity to winter and early spring temperatures, the timing of cherry blossoms is an ideal indicator of the impacts of climate change on tree phenology. Thus, our result might give insight into developing adaptation strategies to climate change in horticulture, conservation planning, restoration, and other related disciplines. In practice, our model could provide tourism guidance (more manageable schedules), pollen season alert, and possibly inspire agricultural planting and induce financial benefits. 

\section{Data}
Our dataset is twofold: Full-flowering (\textgreater{70}\%) date and historical series of phenological data. Both raw data are expected to be consecutive time-sequential, and we would select the intersection dates. Our target regions are Washington D.C. and Kyoto, two cities
renowned for their amazing cherry blossom festival and have comparable geographical features (similar latitude, coastal). For Kyoto, the flowering date data is provided by Yasuki Aono from Osaka Prefecture University, which records the vegetative cycle of the local cherry tree since 810AC \cite{kyoto_flowering}. We would select the 1881-now span as the ancient temperature data is missing. For D.C., the data source is the United States Environmental Protection Agency with records from 1921-2016 for the main type of cherry tree around the Tidal Basin  \cite{dc_flowering}. These peak bloom date data will serve as labels for our classification algorithm. Furthermore, the detailed historical temperature data are from the Japan Meteorological Agency and the U.S. National Oceanic and Atmospheric Administration.  The latter includes multiple daily weather features, like humidity, precipitation, evapotranspiration, and wind speed. However, these data are lacking on the Kyoto side. Such shortage restricts the performance of our model in the following. Our primary preprocessing is cleaning missing values and changing data format. For instance, we modify the original presentation of the date "Month-Day-Year" (timestamp type) to "Date of the Year" (int type), thus eliminating the potential error of leap years. If the average temperature (reported by the measuring station) is missing, we calculate the average of the maximum and minimum temperature on that specific day and fill in the value. 

\section{Methodology}

To select and implement the most suitable model, the first step is to decide whether we should make it a multi-class classification problem or a regression problem.

On one hand, it is quite intuitive to interpret it as a regression problem: for each date, we only need to output a number $n$ indicating the number of days between the date for prediction and the estimated full flowering date, and $n$ could be any value greater than $0$. 

On the other hand, we could also interpret this prediction problem as a multi-class classification problem, in which we will only focus on the peak blossom date estimation within $k$ days. In this case, the output would be a vector with length $k + 1$ containing the probability of class $0$ to class $k$. Class $0$ represents that the estimated peak blossom date is more than $k$ days away, while for $i \in {1, 2, \cdots ,k}$, class $i$ represents that the estimated peak blossom date is $i$ day(s) away.

We finally decided to make it a multi-class classification problem based on the consideration that the multi-class classification approach focuses on a relatively short time span (e.g. $10$ days or $20$ days) and thus could provide more accurate predictions. Although using this method we are not able to predict the full flowering date if it is more than $k$ days away, in this case, the prediction of peak blossom date too far away would be neither valuable nor accurate.

We will implement two different types of models: Support Vector Machine (SVM) classifier and Long Short-Term Memory (LSTM) model to conduct the multi-class classification tasks.

\subsection{Support Vector Machine approach}

Multi-class Suppor Vector Machine (SVM) is essentially a combination of many binary SVM classifiers. Meantime, One-vs-One (OVO) and One-vs-Rest (OVR) are two common methods used to build multiple classification SVM. In our problem, we explicitly choose the OVO scheme to construct our multi-class SVC for two reasons \cite{pawara}. 

\begin{figure}[htb]
\centering
\includegraphics[width=0.4\textwidth]{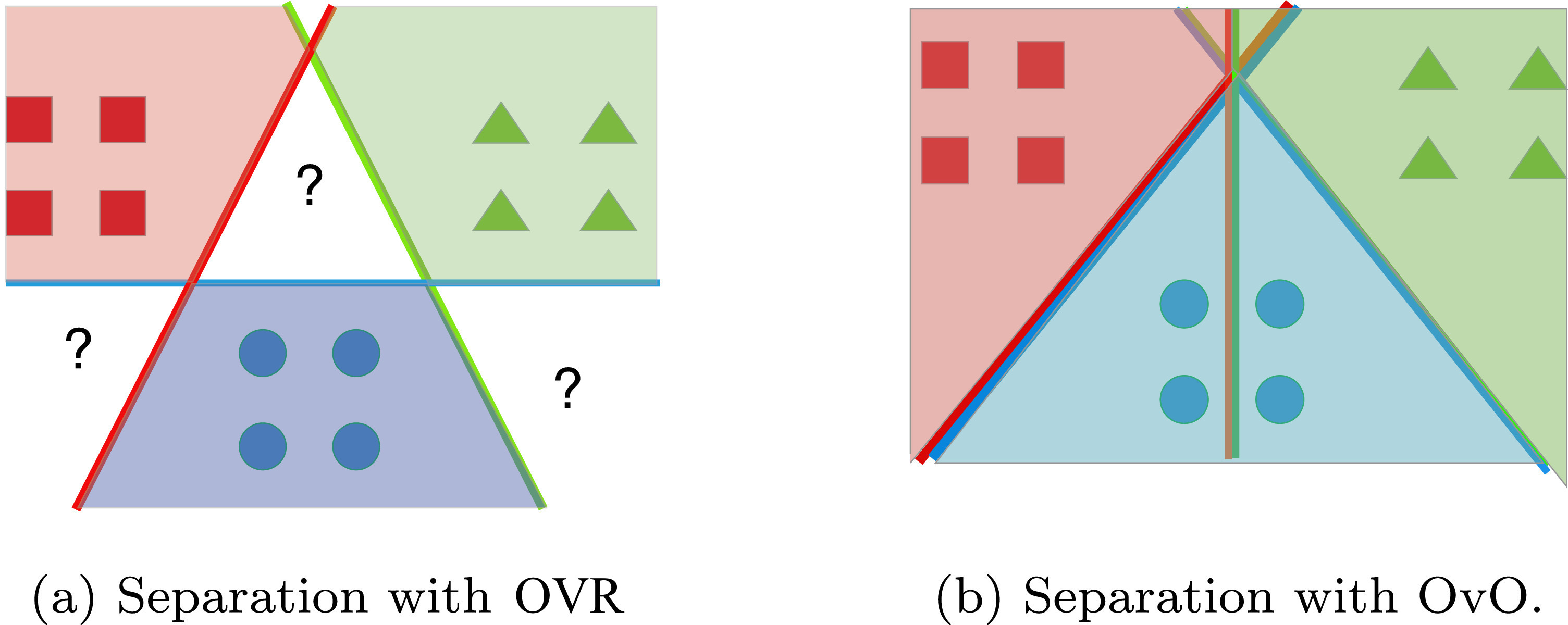}
\caption{\hbox{OVR-OVO schemes}}
\label{fig1}
\end{figure}

First, multi-class classifiers using the OVO scheme do not generate ambiguous regions that further enlarge the bias in the final prediction, and this bias is initially resulted from our imbalanced train dataset. Specifically, in Figure \ref{fig1}, the separation region of the OVR multi-class classifier fails to cover the whole space of data. If an input data $(X_{i}, y_{i})$ lies in the white ambiguous region marked, the OVR-SVC will be confused and will pick a random class near $(X_{i}, y_{i})$ to be the output instead of choosing the one with the largest probability. This kind of prediction is highly susceptible to misclassification in our problem. Because using unbalanced data for training, the generated SVC inevitably favors predicting the majority class appearing most frequently, more accurately than the minority class appearing least frequent \cite{data_mining}. In other words, ambiguous regions import more unfair errors, making the classification results more imbalanced in our multi-class SVC \cite{pawara}. Hence, we chose the OVO scheme over OVR to obtain SVC predicting labels more precisely.

Secondly, classifiers in the OVO scheme are more stable and independent than those in the OVR scheme, “dependent binary classifiers could increase learning instability” \cite{pawara}. Ill-conditioned systems are always unwanted, therefore, we naturally prefer OVO over OVR.

Given $l$ training data $(x_1,y_1),...,(x_l,y_l)$, where $x_0 \in R^n, i = 1,...,l$ and $y_i \in \{ 0,...,10\}$ is the class of $x_i$. The primal problem for each binary soft-margin classifier in our multi-class SVM is: 
\begin{align*}
    \min _{\boldsymbol{w}, b, \boldsymbol{\xi}} & \frac{1}{2} \boldsymbol{w}^{T} \boldsymbol{w}+C \sum_{i=1}^{l} \xi_{i} \\
    \text { subject to } & y_{i}\left(\boldsymbol{w}^{T} \phi\left(\boldsymbol{x}_{i}\right)+b\right) \geq 1-\xi_{i}, \\
    & \xi_{i} \geq 0, i=1, \ldots, l,
\end{align*}
in which, $\frac{1}{2} \boldsymbol{w}^{T} \boldsymbol{w}$ is the margin maximizer, $C \cdot \sum_{i=1}^{l} \xi_{i}$ is the penalty term, $\xi_{i}$ is the slack variable, $C$ is the penalization parameter controlling tolerance of $\xi_{i}$. Since the OVO approach is applied, this model will generate $\frac{k(k-1)}{2}$ sub-classifiers in total, each of them gives us a decision boundary function $f_{i} = \omega^{T} \cdot \phi(x_{i} + b) $. 

Here, because our data is not linearly separable,  we need to transform the feature space, making it separable in other dimensions. So, we apply the RBF kernel trick to complete the transformation. In specific, we choose RBF rather than Linear or polynomial kernel mainly because it generates more flexible boundaries. The Gaussian Radial Basis kernel function is $ exp (\gamma \cdot \lVert{x-\bar{x}}\lVert)$

The final output of the eventual SVM model is: $\space arg\space max (f_{i})\space$, indicating the class $y_{i}$ receiving most votes from $\frac{k(k-1)}{2}$ sub-classifiers will be the final output of our multi-class SVM model. \ref{fig2} shows the detailed flow of our SVM method: 
\begin{figure}[htb]
\centering
\includegraphics[width=0.5\textwidth]{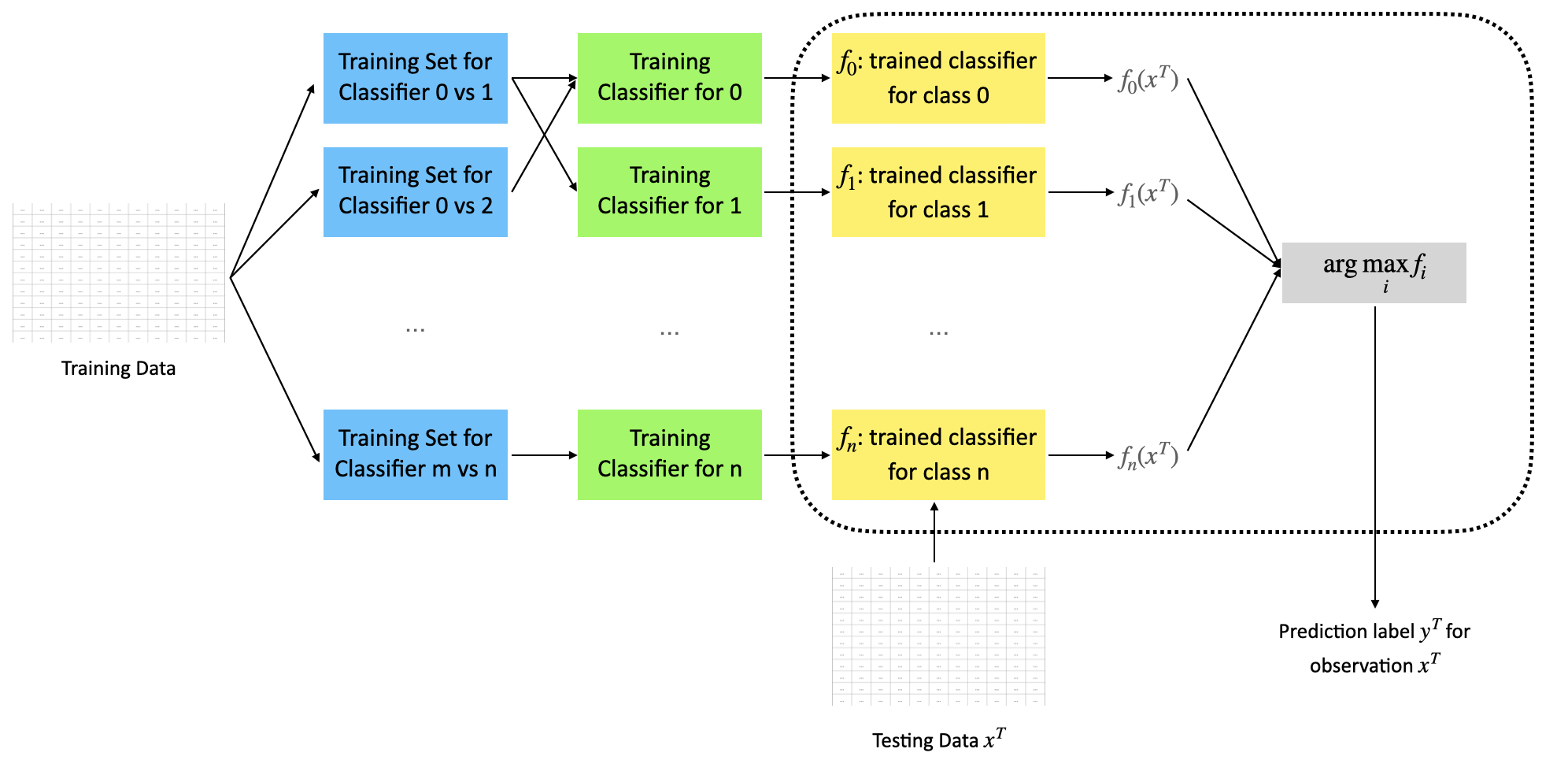}
\caption{\hbox{Multi-class SVC Flow Diagram}}
\label{fig2}
\end{figure}

Yet, the preparation is not done. Recalling our dataset is highly imbalanced, if we left the imbalance problem unsolved and directly do the train-test-split to train and test models, the SVM-classier obtained will be meaningless since it will always generate high accuracy due to its preference for majority class, but fail to be generalized for the minority class. We come up with two approaches to reduce the influence of imbalanced data in our SVC \cite{data_mining}. 

The first approach is to alternate weights of penalization parameter $C_{i}$ of different classes proportionally in the primal equation according to rules: $\space w_{j} = \frac{n}{k\cdot n_{j}} = \hbox{weight} \space \rightarrow \space  C_{j} = C \cdot w_{j}$ ; The second one is to over-sample the training data proportionally of each individual minority class. 

In principle, both approaches are supposed to change the weights of penalization parameters $C^{+}$ and $C^{-}$ in each binary classifier, formulated in the below equation \cite{libsvm},
\begin{align*}
\min _{\boldsymbol{w}, b, \boldsymbol{\xi}} & \frac{1}{2} \boldsymbol{w}^{T} \boldsymbol{w}+C^{+} \sum_{y_{i}=1} \xi_{i}+C^{-} \sum_{y_{i}=-1} \xi_{i} 
\end{align*}

They just achieve the goal in different ways. The first approach directly tunes the proportion of weight in each binary classifier through the built-in \textit{sci-kit-learn} function, while the second change penalization weight through oversampling the size of the targeted minority class through package \textit{imblearn (SMOTE)}. 

Later in the result report, we will compare the evaluation of three circumstances:
\begin{itemize}
\item \textbf{Ordinary SVC} -- Penalizations applied to the majority and minority classes are equal 
\item \textbf{Weighted SVC} -- Penalizations applied to the majority and minority classes are tuned proportionally by the built-in formula in sklearn.
\item \textbf{Oversampled SVC} -- The distributions of input training data are scaled by oversampling individually on each minority class, in our case are the labels $ y_{i}= [1, 2, 3, 4, 5, 6, 7, 8, 9, 10]$. 
\end{itemize}

Then, we will discuss which obtained classifier might be the best to solve our problem in predicting the future flowering of cherry trees. 

\subsection{Long Short-term Memory approach}
Since the temperature data we used is inherently sequential, it is intuitive to think that model that is designed to process sequential data, in particular LSTM, would be a suitable choice. We have two possible approaches available to train the LSTM module, the first one is the many-to-many approach: the model takes the temperature records from all dates before the full flowering date in a particular year as the input and outputs a sequence of predictions for each date. The second one is the many-to-one approach: for temperature records of each year, we use the sliding window approach to generate multiple temperature record sequences of a fixed length $n$ for each year, then the module takes this fixed length sequence as input and outputs a final prediction. For example, to predict the number of days between the full flowering date and March 20, we input the temperature data from March 11 to March 20 and expect the LSTM module to output a prediction (a value for the regression module and a class for classification module) to indicate the number of days between the current date (March 20) to the estimated peak flowering date.

We decided to choose the many-to-one approach based on the following two considerations: first, the input sequence length for the many-to-one approach is relatively shorter, which makes the model simpler and reduces the training time. Secondly, under the many-to-one approach, the input sequence has a fixed length, which facilitates the model implementation. In our model, features such as multi-layer structure and dropout are introduced to improve the prediction accuracy and prevent over-fitting.

\begin{figure}[htb]
\centering
\includegraphics[width=0.45\textwidth]{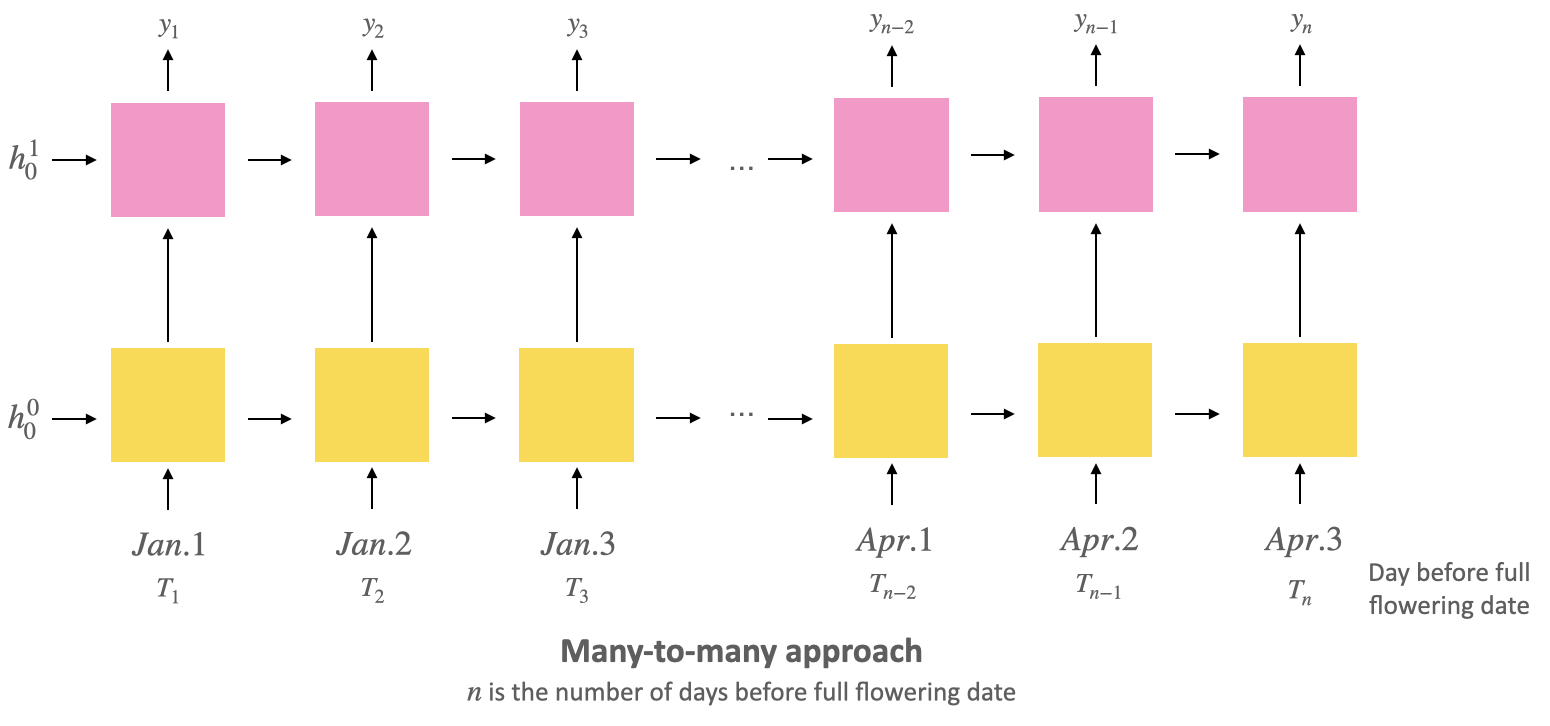}
\caption{\hbox{Many-to-many LSTM Model Structure}}
\end{figure}

\begin{figure}[htb]
\centering
\includegraphics[width=0.45\textwidth]{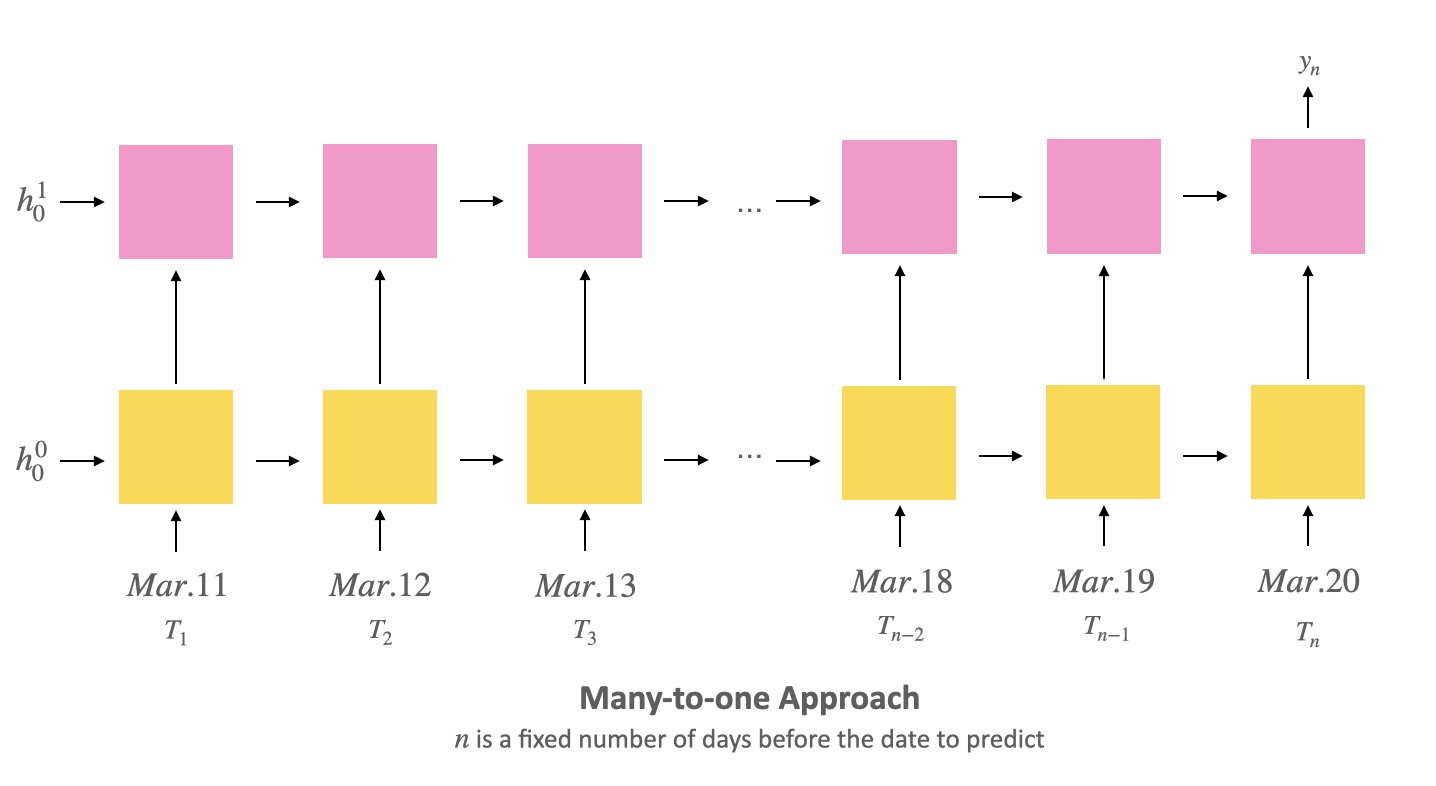}
\caption{\hbox{Many-to-one LSTM Model Structure}}
\end{figure}

The output from the LSTM module is then passed to a dense layer to obtain a vector with a length equal to the number of classes. Finally, this vector is passed to a softmax layer to obtain the probability for each class. The final prediction could be obtained by finding the maximum probability in the output vector.

\section{Results}
\subsection{Oridinary, Weighted, Oversampled muti-class SVC }

In muti-class SVC, we evaluate the models based on Accuracy, Precision, Recall, and F1-score, and visualize them through confusion matrix and PR curves. The results are present in Table \ref{table1}, in which we shall observe as different penalization weights are applied, the accuracy of our Support Vector Classifier decreases greatly as expected. We previously assumed that the high accuracy is resulted from SVC's preference to predict the majority class more accurately while leaving other class predictions with high error. By alternating the penalization effect correspondingly on each class, the preference will be reduced, and the total accuracy shall drop from the perfect level.

$\space$

\begin{table}[htb]
\centering
\begin{tabular}{|c|c|c|c|} 
  \hline
  Models      & Ordinary (clf)  & Weighted (wclf)  & Oversample (oclf)  \\
  \hline
  Accuracy  & 92.86   & 94.27    & 97.89 \\ 
  \hline
  Precision  & 0.435   & 0.783    & 0.998 \\
  \hline
  Recall       & 0.934   & 0.712    & 0.858 \\
  \hline
  F1 Score   & 0.569   & 0.745    & 0.921 \\
  \hline
\end{tabular}
\caption{Evaluation Metrics Comparison (Features set: 10-days-Temperature)}
\label{table1}
\end{table}

\begin{figure}[H]
\centering
\subfigure[Ordinary SVC (clf)]{\label{Fig.sub.1}
\includegraphics[width=0.4\textwidth]{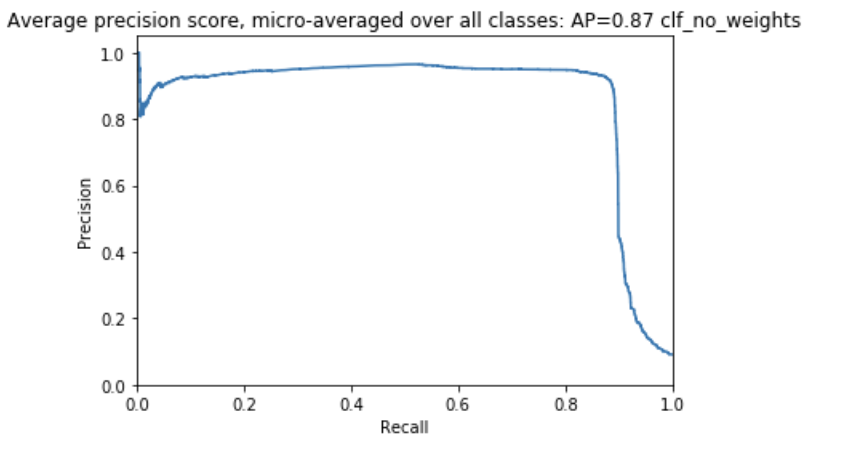}}
\subfigure[Weighted SVC (wclf)]{\label{Fig.sub.2}
\includegraphics[width=0.4\textwidth]{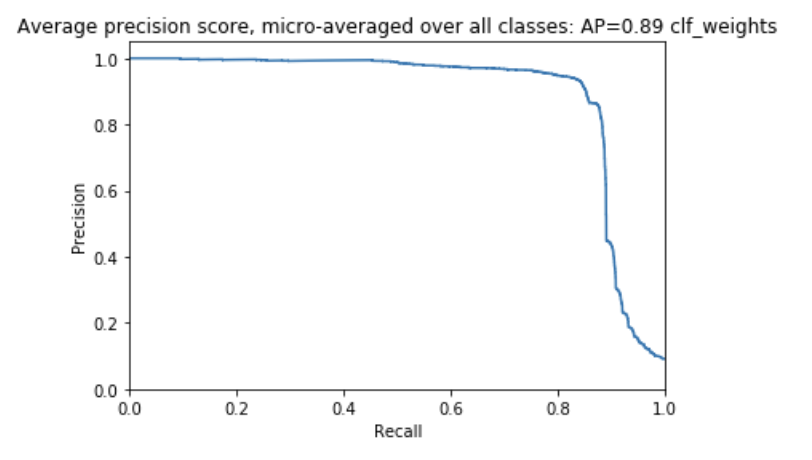}}
\subfigure[Oversampled SVC (oclf)]{\label{Fig.sub.3}
\includegraphics[width=0.4\textwidth]{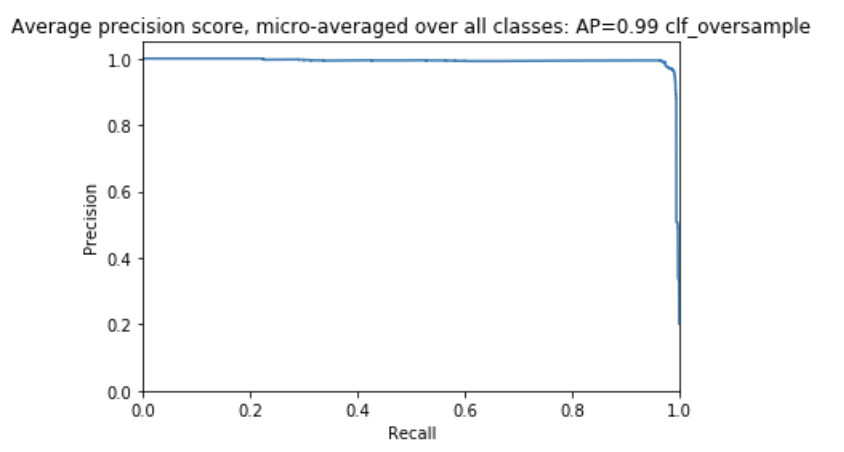}}
\caption{PR-Curves Comparison}
\label{fig6}
\end{figure}

\begin{figure}[H]
\centering
\subfigure[Ordinary SVC (clf)]{\label{Fig.sub.4}
\includegraphics[width=0.4\textwidth]{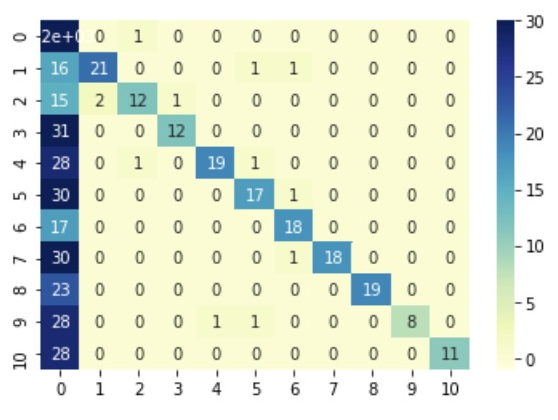}}
\subfigure[Weighted SVC (wclf)]{\label{Fig.sub.5}
\includegraphics[width=0.4\textwidth]{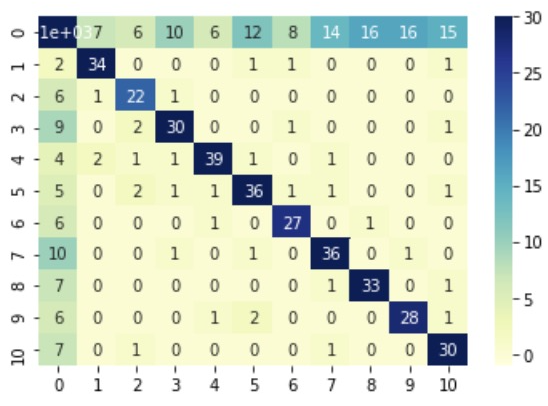}}
\subfigure[Oversampled SVC (oclf)]{\label{Fig.sub.6}
\includegraphics[width=0.4\textwidth]{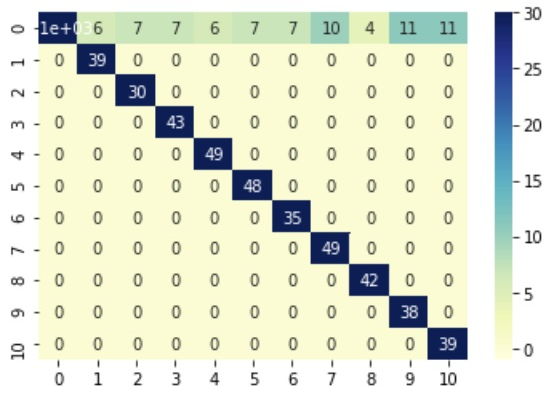}}
\caption{Confusion Matrix Comparison}
\label{Fig.lable}
\end{figure}

Yet, accuracy is not the metric we plan to focus on analyzing. Since our training dataset is highly imbalanced, the value of accuracy is a relatively unreliable evaluation metric compared to the other metrics. To illustrate the effectiveness of weighted penalization, we analyze the Recall and Precision value. From Table \ref{table1}, we shall see that the precision of the model classification result dramatically increases from 0.435 to 0.998, while the recall still keeps at a high level even after diminishing. 

To further understand these changes, one could observe the PR curves changing through the weight-changing process in Figure \ref{fig6}. We can easily interpret that the oversampled SVC performs the best regarding its PR-AUC metric, which is approximately equal to 1, which is nearly perfect. Meanwhile, for the confusion matrix, each cell contains a number indicating the number of correctly classified data pairs of $m \ vs \ n$ (row class v.s column class) classifier. The darker the blue, the better the classification result. As the weighted penalization is applied, the diagonal of the confusion matrix increasingly becomes darker, therefore representing the classification result becoming more and more idealized.

In general, we may confirm the theoretical approach -- applying various penalization -- to solve the imbalance problem in SVM is quite effective in our case, and the resulting multi-class SVC after oversampling is the most accurate one. Yet, we still are not confident with this "nearly perfect" multi-class SVC. To start with, the design of input feature $X_{i}$ = 10-days-temperature might bring over-fitting pitfalls to our classifiers. So we variate the input feature data to another form $M_{i}$ = statistics such as minimum, maximum, average, or variance of 10 days temperature before the input interested date. However, the result becomes chaotic as Table \ref{table2} illustrates. Using $M_{i}$ as the input feature, weighted penalization eventually contradicts our expectation, the precision doesn't increase, and the recall decrease dramatically. Therefore, the final multi-class SVC is useless, and we can hardly be very confident about the model's generalizing ability. 

\begin{table}[h!]
\centering
\begin{tabular}{|c|c|c|c|} 
  \hline
  Models    & Ordinary (clf)  & Weighted (wclf)  & Oversample (oclf)  \\
  \hline
  Accuracy  & 89.68   & 70.87    & 69.16 \\ 
  \hline
  Precision & 0.096   & 0.12    & 0.11 \\
  \hline
  Recall    & 0.724   & 0.112    & 0.106 \\
  \hline
  F1 Score  & 0.094   & 0.110    & 0.101 \\
  \hline
\end{tabular}
\caption{Evaluation Comparison (Features set: Temperature distribution Statistics)}
\label{table2}
\end{table}

Meanwhile, in reality, sequential observations of 10 days' temperature before the interested date are not the only factor determining the blossom of cherry flowers. Sequential daily humidity and wind fluctuations are also influential. Not to mention for cities in different latitudes and longitudes, the city of Kyoto near the sea must be exposed to different factors that motivate yearly full flowering dates from Washington. D.C in the inner land area. Failing to comprehensively include as many weathers and climate features as possible in model training, the predictive effect of our SVM multi-classification model is inevitably limited.

\subsection{LSTM Neural Network}
Our LSTM module is evaluated based on Precision, Recall, and F1-score. The model generates a satisfactory result:

\begin{table}[h!]
\centering
\begin{tabular}{|c|c|} 
  \hline
  Evaluation Metric & Value \\
  \hline
  Loss & 2.31 \\ 
  \hline
  Precision & 19.93 \\
  \hline
  Recall & 13.91 \\
  \hline
  F1 Score & 14.36 \\
  \hline
\end{tabular}
\caption{LSTM Evaluation}
\label{table:1}
\end{table}

\begin{table}[h!]
\centering
\begin{tabular}{|c|c|} 
  \hline
  Parameter & Value \\
  \hline
  num\textunderscore layer & 2 \\ 
  \hline
  input\textunderscore size & 20 \\
  \hline
  hidden\textunderscore size & 30 \\
  \hline
  dropout & 0.5 \\
  \hline
\end{tabular}
\caption{LSTM Module Parameters}
\label{table:2}
\end{table}

Similar to the input data used by the SVM model, the input data for our LSTM module is also based on the data processed by the imblearn package. Since training a deep neural network is quite time-consuming, it is not possible to generate a PR curve.

Although the values of those evaluation metrics may make the impression that our model is not performing well, an important fact that should not be ignored is that this is a multi-class classification problem, and the precision score is much higher than it would be if we simply guess randomly ($9\%$). Besides, since we made it a multi-class classification problem, if the actual peak blossom date is three days away but the prediction is $2$ or $4$, it would be considered an incorrect prediction. Nevertheless, such predictions are quite reasonable as the estimated full flowering date in this case is only one day away from the actual full flowering date. Therefore, it is fair to conclude that the model performance is actually better than those metrics have reflected.

\section{Conclusion}
In general, by designing the feature dataset $X_{i}$ and Labels set $y_{i}$ properly, we could obtain a multi-class SVC and an LSTM recurrent neural network to solve our problem in predicting the specific date of flowering.  Based on the evaluation metrics, we unexpectedly observed that the SVM generally performs better than LSTM in days-of-flowering prediction.  Yet still, since the SVM is constructed upon an imbalanced dataset of [10-days-temperature, class labels] = [$X_{i}, y_{i}$], the eventual obtained multi-class SVC might be highly unreliable in the real case. Therefore, for future works, we might dive deeper into the management of imbalanced datasets in classification, for it is also a common problem in most machine learning research like tumor detection. And if the meteorological data of Kyoto could be found, we would also design the feature data differently in SVM  training to formulate a better multi-class Support Vectors Classifier (SVC).


\end{document}